\documentclass[11pt]{article}

\usepackage[preprint]{acl}

\usepackage{times}
\usepackage{latexsym}

\usepackage[T1]{fontenc}

\usepackage[utf8]{inputenc}

\usepackage{microtype}

\usepackage{inconsolata}

\usepackage{xcolor} 
\definecolor{LightGray}{gray}{0.9}
\usepackage[frozencache=true,cachedir=./minted-cache]{minted}

\usepackage{multirow}

\title{DaVinci at SemEval-2024 Task 9: Few-shot prompting GPT-3.5 for Unconventional Reasoning}

\author{Suyash Vardhan Mathur\thanks{Equal Contribution} \\
  IIIT Hyderabad \\
  {\footnotesize \texttt{suyash.mathur@research.iiit.ac.in\ } } \\\And
  Akshett Rai Jindal\footnotemark[1] \\
  IIIT Hyderabad \\
  {\footnotesize \texttt{\ akshett.jindal@research.iiit.ac.in\ } } \\\And
  Manish Shrivastava \\
  IIIT Hyderabad \\
  {\footnotesize \texttt{\ m.shrivastava@iiit.ac.in} } }

\begin{document}
\maketitle
\begin{abstract}
While significant work has been done in the field of NLP on vertical thinking, which involves primarily logical thinking, little work has been done towards lateral thinking, which involves looking at problems from an unconventional perspective and defying existing conceptions and notions. Towards this direction, SemEval 2024 introduces the task of BRAINTEASER, which involves two types of questions -- Sentence Puzzles and Word Puzzles that defy conventional common-sense reasoning and constraints. In this paper, we tackle both types of questions using few-shot prompting on GPT-3.5 and gain insights regarding the difference in the nature of the two types. Our prompting strategy placed us 26\textsuperscript{th} on the leaderboard for the Sentence Puzzle and 15\textsuperscript{th} on the Word Puzzle task.
\end{abstract}

\section{Introduction}

The human brain consists of two hemispheres - left and right. Both of them are responsible for different kinds of thinking strategies. The left hemisphere is involved in vertical thinking, and the right hemisphere is involved in lateral thinking (\citealp{waks1997lateral}). Vertical (linear, convergent, logical) thinking is a more sequential analytical process. In contrast, in Lateral (outside the box, divergent, creative) thinking, we look at the problem from a new point of view, ignoring the expected associations with items.

In the field of NLP, much research has been done around vertical thinking and significant progress has been made. The recent work around Large Language Models (LLMs) (\citealp{devlin2018bert}; \citealp{openai2022}) has achieved great performance in solving complex reasoning tasks (\citealp{talmor2018commonsenseqa}; \citealp{bisk2020piqa}; \citealp{sap-etal-2019-social}). This performance is consistent in both cases when no task examples have been provided to the model during inference (zero-shot) (\citealp{sanh2022multitask}) and when the model is introduced with the task during inference time (few-shot) (\citealp{chung2022scaling}).

However, lateral thinking has been overlooked when training NLP models like LLMs. When creating datasets for various models, texts that involve lateral thinking are mostly considered noise and filtered out from the data because researchers want their models to perform better at traditional reasoning tasks and not get confused by lateral thinking.

The task {\Large B}RAIN{\Large T}EASER (\citealp{jiang-etal-2023-brainteaser}, \citealp{jiang-semeval-2024-brainteaser}) tries to bridge this gap that exists between vertical and lateral thinking for LLMs and other NLP models. They formulated a set of Multi-choice Question Answers containing puzzles that can be solved only using lateral thinking. The benchmark dataset contains two types of lateral thinking puzzles - Sentence Puzzles and Word Puzzles. This has been constructed by designing a data collection procedure that crawled relevant puzzles from many websites that were publicly available performing semi-automatic filtering of irrelevant questions.

\section{Background}

\subsection{Dataset}

The dataset being used in this task is {\Large B}RAIN{\Large T}EASER (\citealp{jiang-etal-2023-brainteaser}). It was prepared by scraping puzzles from various publicly available websites and then semi-automatically filtering them out. Then \textit{semantic reconstruction} and \textit{context reconstruction} techniques were used to create variants of each puzzle without affecting its out-of-the-box thinking style. This helped in preventing possible memorization by LLMs and the lack of consistency of the puzzles.

The puzzles in this dataset can be divided into two categories:
\begin{itemize}
    \item \textbf{Sentence Puzzles}: These are brain teasers where the puzzle-defying commonsense is centered on sentence snippets.

    For example, \textbf{Question}: \textit{You are running so fast but you're not getting closer. Where are you?} \textbf{Answer}: \textit{Treadmill.} \textbf{Explanation}: This is because while running on a treadmill, we stay put where we are. The key is understanding that running on a treadmill means you remain stationary despite the motion.

    \item \textbf{Word Puzzles}: These are brain teasers where the answer violates the default meaning of the word and focuses more on the letter composition.

    For example, \textbf{Question}: \textit{How can you make "ten" out of "net"?} \textbf{Answer}: \textit{Just flip it around.} \textbf{Explanation}: This is because if we consider the spelling of the word "ten" and we flip the letters of the word around, we get the word "net" which is what we want to make out of "ten".
\end{itemize}

The training data contains \textbf{507 Sentence Puzzles} and \textbf{396 Word Puzzles}. Each of these puzzles has 4 options to choose from and only one option is the correct answer.

\subsection{Related Works}

With the recent success of LLMs in various NLP tasks, researchers have also started exploring their use for Multiple Choice Question Answering (MCQA) tasks (\citealp{robinson2022leveraging}; \citealp{zheng2023large}).

Researchers have also started employing the technique of \textbf{few-shot prompting} (\citealp{liu2023pre}; \citealp{ma2024fairness}; \citealp{lu2021fantastically}) for various tasks and it has shown improvements when compared with \textbf{zero-shot prompting}.

LLMs like GPT-3.5 have been trained on vast amounts of human-generated text. The main features around which such models are trained are \textbf{Pattern Recognition}, \textbf{Creative Reasoning} and \textbf{Wide Knowledge Range}.

Thus, we decided to employ few-shot prompting on LLMs for this task.

\section{System Overview}
Our architecture uses GPT-3.5 (\citealp{gpt3.5}) (specifically \textit{gpt-3.5-turbo}) with few-shot prompting to answer the question.

\subsection{GPT-3.5}
In NLP, the architecture of Generative Pre-trained Transformer (GPT) 3.5 (GPT-3.5) stands as a significant advancement, which is the culmination of iterative improvement over its predecessors. The architecture of the model is based upon the Transformer model (\citealp{NIPS2017_3f5ee243}), which uses self-attention to enhance performance over the prior sequential models. GPT-3.5 scales this Transformer architecture to over hundereds of billions of parameters, which have been trained by exposing and training the model on hundreds of billions of tokens.

In particular, due to the autoregressive nature of GPT-3.5 and due to being trained on extremely large data, it has enough knowledge about the language and the real world to perform tasks in a Zero-shot setting (\citealp{sanh2022multitask}).  This Zero-shot setting allows the model to understand and execute a task it hasn't been explicitly trained for. These capabilities have been reflected in GPT-3.5 being used in Summarization (\citealp{liu2023abstractive}), Question Answering (\citealp{bahak2023evaluating}), Natural Language Inference (\citealp{ye2023comprehensive}), etc.

\subsection{Few-shot prompting}
While zero-shot prompting works well for simple tasks, tasks like BrainTeaser are a bit more complex in nature, and in such cases providing explicit instructions to the LLM about the nature of the task along with few examples of the task \textit{(few shots)} becomes extremely helpful for the model (\citealp{chung2022scaling}). Here, the few-shot technique involves providing GPT-3.5 some examples, allowing GPT-3.5 to generalize from the few examples, drawing on its large pre-trained knowledge about the language and the real-world.

Thus, 2 different sets of prompts are created for the task, one for the Sentence Puzzle Task and another for the Word Puzzle task, since the 2 tasks are fundamentally different and need different instructions and examples.

\subsection{Experimental Setup}

\begin{listing}[t!]
    \begin{minted}[frame=lines,
                    framesep=3mm,
                    linenos=false,
                    bgcolor=LightGray,
                    fontsize=\tiny,
                    breaklines=true,
                    tabsize=2]{text}
You are given a question with multiple choices that you need to answer. The answer would only be one index of the multiple choices available. Such a question would involve  brain teaser questions where the puzzle defying commonsense is centered on sentence snippets.
IMPORTANT: It's crucial to analyze the question from an unconventional perspective, focusing on the literal or alternative meanings of the words used, rather than relying on common sense. You must not use commonsense, but look at meaning from a different perspective than what would commonly be done. For example,

Example 1:
Question: You are running so fast but you're not getting closer. Where are you?

Option 0: Country road.
Option 1: Treadmill.
Option 2: High way.
Option 3: None of above.

Answer: 1
Reason: This is because while running on a treadmill, we stay put where we are. The key is understanding that running on a treadmill means you remain stationary despite the motion. This is not valid for Country road or High way. Thus, the answer is 1 - Treadmill.

Example 2:
Question: From elementary school to collage, how many "first day of school" does the average person have in their lifetime?

Option 0: They technically only have one first day of school in their lifetime. That's the very first day they started attending school as a child.
Option 1: Average people have 4: elementary school, middle school, high school, and college.
Option 2: Average people have "first day of school" in each semester, so it will be more than 10!
Option 3: None of above.

Answer: 0
Reason: First day of school can only be one day in a person's lifetime. Here, it is important to understand that first day of middle school, high school, college won't be first day of school. Similarly, each semester's first day is not TECHNICALLY first day of school. This, the answer is 0 -  They technically only have one first day of school in their lifetime. That's the very first day they started attending school as a child. Thus, the key here is the term 'first day of school' technically refers to the very first day a person attends school, making all subsequent 'first days' at different educational levels irrelevant to the specific question.

Now, using these examples, answer the question below. It is IMPORTANT that you just provide the index of the answer in the response. DO NOT output the reason behind choosing the answer:

Question: In a small village, two farmers are working in their fields - a diligent farmer and a lazy farmer. The hardworking farmer is the son of the lazy farmer, but the lazy farmer is not the father of the hardworking farmer. Can you explain this unusual relationship?
Option 0: The lazy farmer is his mother.
Option 1: The lazy farmer is not a responsible father as he is lazy.
Option 2: The diligent farmer devoted himself to the farm and gradually forgot his father.
Option 3: None of above.

Answer:
    \end{minted}
    \caption{Prompt for the Sentence Puzzle}
    \label{listing:sample-conversation}
\end{listing}

\begin{listing}[t!]
    \begin{minted}[frame=lines,
                    framesep=3mm,
                    linenos=false,
                    bgcolor=LightGray,
                    fontsize=\tiny,
                    breaklines=true,
                    tabsize=2]{text}

You are given a question with multiple choices that you need to answer. The answer would only be one index of the multiple choices available. The question demands an unorthodox approach, focusing on the spellings or structural aspects of words, rather than their standard meanings. Your task is to choose the correct answer from the given multiple-choice options by analyzing the words in a literal or unconventional way.
IMPORTANT: It's crucial to analyze the question from an unconventional perspective, focusing on the spellings of certain words, rather than relying on common sense. You must not use commonsense, but look at meaning from a different perspective considering arrangement of the letters in certain words than what would commonly be done. For example,

Example 1:
Question: How can you make "ten" out of "net"?

Option 0: Just flip it around.
Option 1: Remove the letter "e".
Option 2: Move the letter "t" to the end.
Option 3: None of above.

Answer: 0
Reason: This is because if we consider the spelling of the word 'ten' and we flip the letters of the word 'ten' around, we get the word 'net', which is what we want to make out of 'ten'.  The answer focuses on the literal rearrangement of the letters, disregarding the typical meanings of the words. Thus, the answer is 0 - Just flip it around.

Example 2:
Question: What is the most fast city?

Option 0: Urban city.
Option 1: Inner city.
Option 2: Velocity.
Option 3: None of above.

Answer: 2
Reason: The term 'fast' in the question prompts an unconventional interpretation. All options contain the word "city", but "velocity" stands out as it directly relates to speed or 'fastness'. The question cleverly uses the term 'city' as a red herring, while the actual focus is on the concept of speed.

Now, using these examples, answer the question below. It is IMPORTANT that you just provide the index of the answer in the response. DO NOT output the reason behind choosing the answer:

Question: What sort of cheese is made in reverse?
Option 0: Cheddar cheese..
Option 1: Edam cheese.
Option 2: Blue cheese.
Option 3: None of above.

Answer:
\end{minted}
    \caption{Prompt for the Word Puzzle}
    \label{listing:word-puzzle prompt}
\end{listing}

We provide 2-shot prompts to GPT-3.5 for our leaderboard submission. We also try out 5-shot prompt in the post evaluation phase to test if providing more examples helps the model perform better.

The prompt used for the Sentence Puzzle task is shown in Listing \ref{listing:sample-conversation}. As we can see, the prompt first details the task, and \textit{IMPORTANT} keyword is used to express to GPT-3.5 that commonsense must not be used in the task, but instead it should look at meaning from an unconventional sense. Then, 2 examples are given, along with \textbf{reasoning} behind the answers too. This was important, as this gave the model more knowledge to be able to generalize the task from the examples. Further, the output format was clearly specified in the prompt so as to avoid getting extra information in the model output.

Similar prompt for Word Puzzle can be seen in Listing \ref{listing:word-puzzle prompt}. The prompt clarifies that the structural aspect of the words should be focused on, emphasizing that unconventional meaning should be looked at. Then, 2 examples that exhibit structural aspect are given along with reasoning behind their answers as well as constraints for the output format.

\section{Results and Analysis}

\begin{table*}[!t]
  \begin{center}
  \small
    \caption{Results of zero-shot and few-shot prompting on GPT-3.5 for the two \textsc{BrainTeaser} subtasks. Ori = Original, Sem = Semantic, Con = Context.}

    \label{tab:few-shot}
   \begin{tabular}{|c|ccc|cc|c|}
   \hline
          \multirow{2}{*}{\textbf{Model}} &   \multicolumn{3}{|c|}{\textbf{Instance-based}} & \multicolumn{2}{|c|}{\textbf{Group-based}} & \multirow{2}{*}{\textbf{Overall}} \\
     \cline{2-6}
       &  \textbf{Original}     &    \textbf{Semantic}   &    \textbf{Context}  &  \textbf{Ori \& Sem }  &  \textbf{Ori \& Sem \& Con }  &         \\      \hline
      \multicolumn{7}{c}{\textbf{\textit{Sentence puzzle}}}           \\
   \hline
        GPT-3.5 (zero-shot) \textbf{Baseline} & 60.7 & 59.3 & \textbf{67.9} & 50.7 & 39.7 & \textbf{62.6} \\ 
        GPT-3.5 (two-shot)  & 57.5 & 55.0 & 42.5 & 50.0 & 30.0 & 51.7   \\ 
        GPT-3.5 (five-shot)  & \textbf{62.5} & \textbf{65.0} & 55.0 & \textbf{62.5} & \textbf{42.5} & 60.8   \\ 

    \hline
         \multicolumn{7}{c}{\textbf{\textit{Word puzzle}}}           \\
   \hline
        GPT-3.5 (zero-shot) \textbf{Baseline} & 56.1 & 52.4 & 51.8 & 43.90 & 29.3 & 53.5   \\ 
        GPT-3.5 (two-shot)& 71.9 & 71.9 & 62.5 & 59.4 & 46.9 & 68.6   \\ 
        GPT-3.5 (five-shot)& \textbf{78.1} & \textbf{90.6} & \textbf{84.4} & \textbf{78.1} & \textbf{68.8} & \textbf{84.4}   \\ 
   \hline
    \end{tabular}
  \end{center}
\end{table*}

The results are detailed in Tab.\ref{tab:few-shot}. For comparison, we also list the zero-shot prompting results reported in \citealp{jiang-etal-2023-brainteaser}. As we can see, the two-shot performance on Word puzzle improved over the zero-shot setting for all the categories, while the same worsened in case of the Sentence puzzle.

This is because of the very nature of the two problems. Sentence puzzle involves deeper non-conventional semantic understanding of the question and the choices, which despite conveying reasoning behind the answers in the few-shot examples, cannot be generalized as easily with just 2 examples. On the other hand, the only tricky component of the Word Puzzle is that the structural aspect of certain words needs to be taken instead of the actual surface meaning of the said words. This can be much more easily generalized through just as few as two examples in the prompt. Further, adding the examples in the Sentence puzzle that don't generalize very well for other questions in the testing set might have acted as noise for the model, which led to poorer performance.

We also note that using five-shot prompt instead of two-shot prompt hugely increases the performance. This is to be expected, as providing more examples would help the model generalize even better towards solving the task. This is specially true for Word Puzzle questions, where adding more examples allows the model to generalize the task much better.

However, in Sentence Puzzle we still notice a drop in the overall performance as compared to the zero-shot model. This is because of a drop in the performance of the context reconstruction questions, and a marginal increase in comparison to zero-shot in other types of questions. However, group based accuracy increases in five-shot, which might indicate that with five examples, the model is able to handle the variations in reconstructions better, albeit with performance of Contextual Reconstruction taking a dip. These observations are in line with the drop observed in two-shot prompt in comparison to the zero-shot prompt, highlighting the difficult nature of the task of Sentence Puzzle questions and the inability of the model to generalize using few Sentence Puzzle examples. However, we do note that the performance on Sentence Puzzle also does improve with additional examples between two-shot and five-shot prompting.
\section{Conclusion}
In conclusion, we explored the effectiveness of few-shot prompting for LLMs for complex and unconventional tasks. Further, it demonstrates that few-shot prompting is helpful only in scenarios where the examples convey enough information that can be better generalized, as the results worsened in the Sentence Puzzle while improved in the Word Puzzle.

In future, better prompting strategies like Chain of Thought prompting (\citealp{wang-etal-2023-towards}) can be utilized to improve the performance. Additionally, finetuning the pre-trained LLMs might also help in the task further. Also, increasing the number of training examples might help in further improving the model performance, as observed in the gains of performance in the five-shot prompt in comparison to the two-shot prompt.

\bibliography{custom}

\end{document}